\documentclass[10pt,twocolumn,letterpaper]{article}

\usepackage{3dv}
\usepackage{times}
\usepackage{epsfig}
\usepackage{graphicx}
\usepackage{amsmath}
\usepackage{amssymb}
\usepackage{multirow}

\usepackage{pdfpages}
 
\usepackage{comment}

\usepackage[font=footnotesize, labelfont=footnotesize, labelfont=bf]{caption}

\usepackage[pagebackref=true,breaklinks=true,letterpaper=true,colorlinks,bookmarks=false]{hyperref}

\def\bfX{{\bf X}} 
\def\bfY{{\bf Y}}

\def\bfx{{\bf x}} 
\def\bfy{{\bf y}} 

\def\bfv{{\bf v}} 
\def\bfV{{\bf V}}

\def\bfw{{\bf w}} 

\def\bfd{{\bf d}} 

\def\bfD{{\bf D}}

\ifthreedvfinal\fi

\newcounter{alphasect}
\def\alphainsection{0}

\let\oldsection=\section
\def\section{%
  \ifnum\alphainsection=1%
    \addtocounter{alphasect}{1}
  \fi%
\oldsection}%

\renewcommand\thesection{%
  \ifnum\alphainsection=1%
    \Alph{alphasect}%
  \else%
    \arabic{section}%
  \fi%
}%

\newenvironment{alphasection}{%
  \ifnum\alphainsection=1%
    \errhelp={Let other blocks end at the beginning of the next block.}
    \errmessage{Nested Alpha section not allowed}
  \fi%
  \setcounter{alphasect}{0}
  \def\alphainsection{1}
}{%
  \setcounter{alphasect}{0}
  \def\alphainsection{0}
}%
\threedvfinalcopy %
\begin{document}
\title{DispVoxNets: Non-Rigid Point Set Alignment with Supervised Learning Proxies\thanks{supported by the ERC Consolidator Grant 4DReply (770784) and the BMBF project VIDETE (01IW18002).}}

\author{Soshi Shimada$^{1,2}$\hspace{0.8em}
Vladislav Golyanik$^3$\hspace{0.8em}
Edgar Tretschk$^3$\hspace{0.8em}
Didier Stricker$^{1,2}$\hspace{0.8em}
Christian Theobalt$^3$\vspace{0.5em}\\
\hspace{-20pt}
$^{1}$University of Kaiserslautern $\;\;\;\;\;\;\;\;\;\;\;\;\;\;$ 
$^{2}$DFKI $\;\;\;\;\;\;\;\;\;\;\;\;\;\;$ 
$^{3}$MPI for Informatics, SIC     
}

\maketitle 

\begin{abstract} 
   We introduce a supervised-learning framework   
   for non-rigid point set alignment of a new kind 
   --- {\normalfont Displacements on Voxels Networks} (\hbox{DispVoxNets}) --- 
   which abstracts 
   away from the point set representation and regresses 3D displacement fields on regularly sampled proxy 3D voxel grids. 
   Thanks to recently released collections of deformable objects with known intra-state correspondences, 
   \hbox{DispVoxNets} learn a deformation model and further priors (\textit{e.g.,}~weak point topology preservation) for different object categories such as cloths, human bodies and faces. 
   \hbox{DispVoxNets} cope with large deformations, noise and clustered outliers more 
   robustly than the state-of-the-art. 
   At test time, our approach runs orders of magnitude faster than previous techniques. 
   All properties of \hbox{DispVoxNets} are ascertained numerically and qualitatively in extensive experiments and comparisons to several previous methods. 
\end{abstract}

\section{Introduction}\label{sec:introduction}

Point sets are raw shape representations which can implicitly encode surfaces and volumetric structures with inhomogeneous sampling densities. 
Many 3D vision techniques generate point sets which need to be subsequently aligned for various tasks such as shape recognition, appearance transfer and shape completion, among others. 

The objective of non-rigid point set registration (NRPSR) is the recovery of a general displacement field aligning \textit{template} and \textit{reference} point sets, as well as correspondences between those. In contrast to rigid or affine alignment, where all template points transform according to a single shared transformation, in NRPSR, every point of the template has an individual transformation. 
Nonetheless, real structures do not evolve arbitrarily and 
often preserve the point topology. %
\begin{figure}[t!] 
 \centering 
  \includegraphics[width=1.0\linewidth]{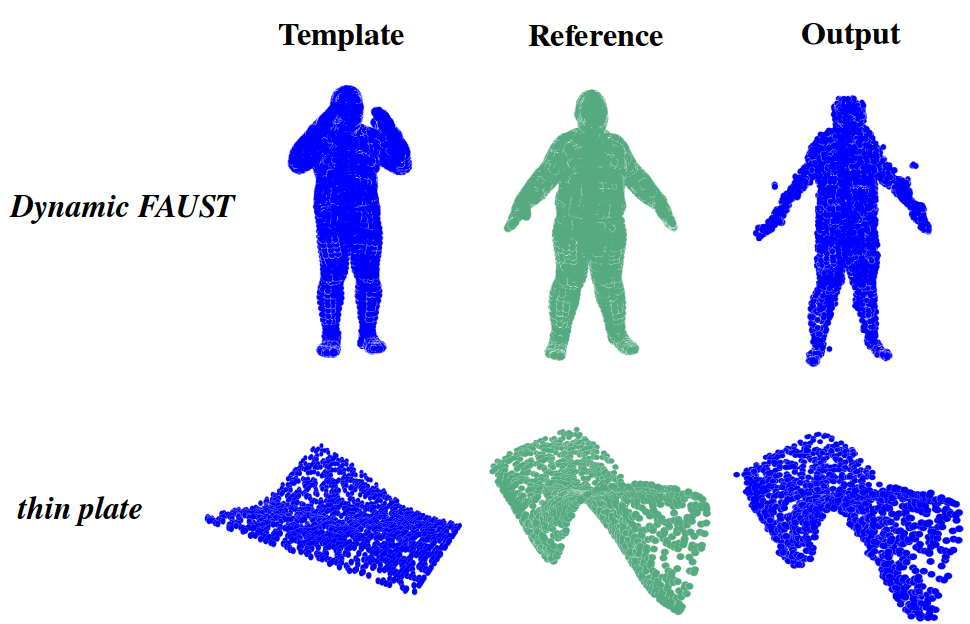} 
  \caption{ 
  Alignment results of \textit{human body scans} \cite{dfaust2017} and \textit{thin plate} \cite{Golyanik2018} with our \hbox{DispVoxNets}. In both cases, the template and reference differ by large non-linear deformations (articulated motion in the case of human body scans). To the best of our belief, \hbox{DispVoxNet} is the first 
  non-rigid point set alignment approach which learns object-specific deformation models purely from data and does not rely on engineered priors. 
  } 
  \label{fig:teaser} 
\end{figure}

\begin{figure*}[t!]
 \centering
  \includegraphics[width=1.0\linewidth]{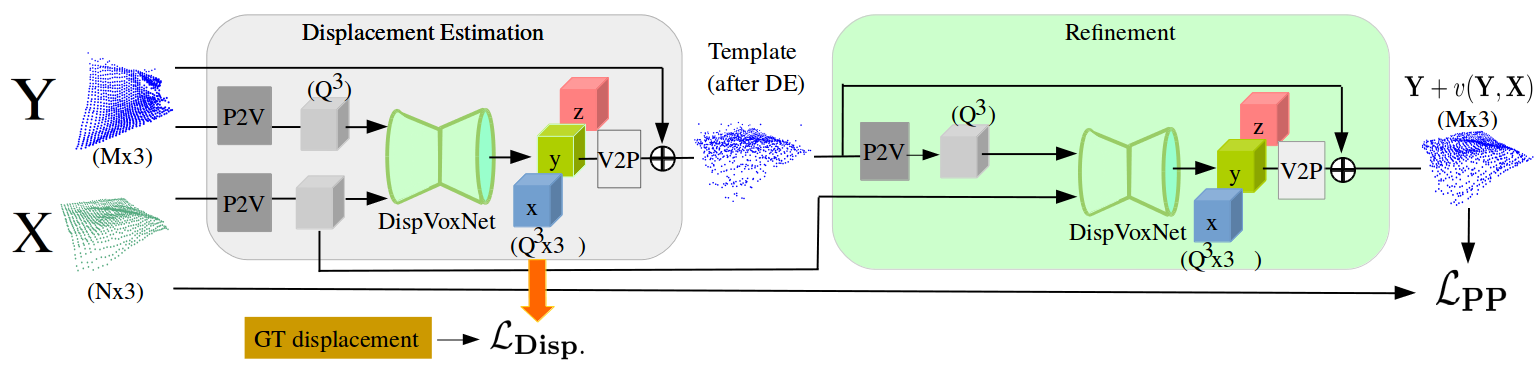} 
  \caption{Overview of our approach. The objective is to non-rigidly align a template $\bfY$ to a reference $\bfX$. In the displacement estimation stage, we first convert the point sets to a voxel representation (P2V). \hbox{DispVoxNets} then regress  per-voxel displacements that we apply to $\bfY$ (V2P), see Table~\ref{tab:architecture_detail} for more details on the network architecture. 
  The first stage is trained in a supervised manner with ground truth displacements using the \textit{displacement loss} ($\mathcal{L}_{\text{\bf{Disp.}}}$). The subsequent stage refines the displacements with  the unsupervised \textit{point projection loss} ($\mathcal{L}_{\text{\bf{PP}}}$). Trilinear weights are applied to the displacements for interpolation and are used to compute weighted gradients in the backward pass, 
  see  Fig.~\ref{fig:grad_table} and %
  our supplement 
  for more details on the trilinear interpolation.} 
  \label{fig:architecture} 
\end{figure*} 

\subsection{Motivation and Contributions} 

On the one hand, existing general-purpose NRPSR techniques struggle to align point clouds differing by large non-linear deformations or articulations (\textit{e.g.,} significantly different facial expressions or body poses) and cause overregularisation, flattening and structure distortions \cite{Besl1992, Chui2003, Myronenko2010, Jian2005}. 
On the other hand, specialised methods exploit additional engineered (often class-specific) priors to align articulated and highly varying structures \cite{Mundermann2007, Pellegrini2008, Ge2014, tagliasacchi2015robust, taylor2016efficient, Golyanik2016}. 
In contrast, \textit{we are interested in a general-purpose method supporting large deformations and articulations (such as those shown in Fig.~\ref{fig:teaser}), which is robust to noise and clustered outliers and which can adapt to various object classes.} 

It is desirable but challenging to combine all these properties into a single technique. 
We address this difficult problem with supervised learning on collections of deformable objects with known intra-state correspondences.
Even though deep learning is broadly and successfully applied to various tasks in computer vision, its applications to NRPSR have not been demonstrated in the literature so far (see Sec.~\ref{sec:related_work}). 
One of the reasons is the varying cardinalities of the inputs, which poses challenges in the network architecture design.
Another reason is that sufficiently comprehensive collections of deformable shapes with large deformations, suitable for the training have just recently become available \cite{dfaust2017, bednarik2018learning, Golyanik2018,FLAME:SiggraphAsia2017}. 

\textit{To take advantage of the latter, our core idea is to associate  deformation priors with point displacements 
and predict feasible category-specific deformations between input samples on an abstraction layer.} 
At its core, our framework contains geometric proxies  --- deep convolutional encoder-decoders 
operating on voxel grids --- which learn a class-specific deformation model. 
We call the proposed proxy component \textit{Displacements on Voxels Network} (\hbox{DispVoxNet}). 
Our architecture contains two identical \hbox{DispVoxNets}, \ie, one for global displacements (trained in a supervised manner) and one for local refinement (trained in an unsupervised manner).  

The proposed \hbox{DispVoxNets} abstract away from low-level properties of point clouds such as point sampling density and ordering. 
They realise a uniform and computationally feasible lower-dimensional  parametrisation of deformations which are eventually transferable to the  template in its original resolution and configuration. 
At the same time, \hbox{DispVoxNets} handle inputs of arbitrary sizes. 
To bridge a possible discrepancy in resolution between the 3D voxel grids and the point clouds, we maintain a point-to-voxel affinity table and apply a super-resolution approach. 
Due to all these properties, \hbox{DispVoxNet} enables the level of generalisability of our architecture which is essential for a general-purpose NRPSR approach. 

A schematic overview of the proposed architecture with \hbox{DispVoxNets} is given in Fig.~\ref{fig:architecture}. 
Our general-purpose NRPSR method can be trained for arbitrary types of deformable objects. 
During inference, no further assumptions about the input point sets except of the object class are made. 
All class-specific priors including the weak topology preserving constraint are learned directly from the data. 
Whereas some methods model noise distributions to enable robustness to noise \cite{Myronenko2010}, we augment the training datasets by adding uniform noises and removing points uniformly at random. 
We do not rely on parametric models, pre-defined templates, landmarks or  known segmentations (see Sec.~\ref{sec:approach}). 

In our experiments, \hbox{DispVoxNets} consistently outperform other tested approaches in scenarios with large deformations, noise and missing data.  
In total, we perform a study on four object types and show that  \hbox{DispVoxNets} can efficiently \textit{learn} class-specific priors (see Sec.~\ref{sec:experiments}).

\section{Related Work}\label{sec:related_work}

\paragraph{Methods with Global Regularisers.} %
When correspondences between points are given, an optimal rigid transformation between the point sets can be estimated in a closed form \cite{Horn1988}. 
Iterative Closest Point (ICP) alternates between estimating the correspondences based on the nearest neighbour rule and local transformations %
until convergence \cite{Besl1992, ChenMedioni1991}. 
ICP is a simple and widely-used point set alignment technique, with multiple policies available to improve its convergence properties, runtime and robustness to noise \cite{GreenspanGodin2001, Rusinkiewicz2001, Granger2002, Fitzgibbon2003}. 
In practice, conditions for a successful alignment with ICP (an accurate initialisation and no disturbing effects such as noise) are often not satisfied. 
Extensions of ICP for the non-rigid case employ thin splines for topology regularisation \cite{Chui2003}\footnote{introduced by Chui \textit{et al.}~\cite{Chui2003} as a baseline ICP modification}
or Markov random fields linked by a non-linear potential function \cite{Hahnel2003}.

Probabilistic approaches operate with multiply-linked point associations. 
Robust Point Matching (RPM) with a thin-plate splines (TPS) deformation model \cite{Chui2003} uses a combination of soft-assign \cite{Gold97} and deterministic annealing for non-rigid alignment. %
As the transformation approaches the optimal solution, 
the correspondences become more and more certain. %
In \cite{Zheng2006}, point set alignment is formulated as a graph matching problem which aims to maximise the number of matched edges in the graphs. 
The otherwise NP-hard combinatorial graph matching problem is approximated as a constrained optimisation problem with continuous variables by relaxation labeling. 
In Gaussian Mixture Model (GMM) Registration (GMR) \cite{Jian2005}, NRPSR is interpreted as minimising the distance between two mixtures of Gaussians with a TPS mapping. 
The method was shown to be more tolerant to outliers and more statistically robust than TPS-RPM and non-rigid ICP. 
Myronenko and Song~\cite{Myronenko2010} interpret NRPSR as fitting a GMM (template points) to data (reference points) and regularise displacement fields using motion coherence theory. 
The resulting Coherent Point Drift (CPD) was shown to handle noisy and partially overlapping data with unprecedented accuracy. 
Zhou \textit{et al.}~\cite{Zhou2014} investigate the advantages of the Student's-t Mixture Model over a GMM. %
The method of Ma \textit{et al.}~\cite{Ma2013} alternates between correspondence estimation with shape context descriptors and transformation estimation by minimising the $\ell_2$ distance between two densities. 
Their method demands deformations to lie in the reproducing kernel Hilbert space.

Recently, physics-based alignment approaches were discovered \cite{Deng2014, Golyanik2016CVPR, Ali2018, Jauer2019, BHRGA2019}. 
Deng \textit{et al.}~\cite{Deng2014} minimise a distance metric between  
Schr\"{o}dinger distance transforms performed on the point sets. %
Their method has shown an improved recall, \textit{i.e.,} the portion of correctly recovered correspondences. 
Ali \textit{et al.}~\cite{Ali2018} align point sets as systems of particles with masses deforming under simulated gravitational forces. 
Gravitational simulation combined with smoothed particle hydrodynamics regularisation place this approach among the most resilient to large amounts of uniform noise in the data, and, at the same time, most computationally expensive techniques. 
In contrast, the proposed approach executes in just a few seconds and is robust to large amounts of noise due to our training policy with noise augmentation.

\paragraph{Large Deformations and Articulations.} 
If point sets differ by large deformations and articulations,  global topology regularisers of the methods discussed so far often overconstrain local deformations. %
Several extended versions of ICP address the case of articulated bodies with the primary applications to human hands and bodies \cite{Mundermann2007, Pellegrini2008, tagliasacchi2015robust}. 
Ge~\textit{et al.}~\cite{Ge2014} extend CPD with a local linear embedding which accounts for multiple non-coherent motions and local deformations. 
The method assumes a uniform sampling density and its accuracy promptly decays with an increasing level of noise. 
Some methods align articulated bodies with problem-specific segmented templates~\cite{Golyanik2017}. 
In contrast to all these techniques, our \hbox{DispVoxNets} can be trained for an arbitrary object class and are not restricted to a single template. 
Furthermore, our approach is resilient to sampling densities, large amounts of outliers and missing data. 
It grasps the intrinsic class-specific deformation model on multiple scales (global and localised deformations) directly from data. 
A more in-depth overview and comparison of NRPSR methods can be found in \cite{Tam2013, Zhu2019}.

\paragraph{Voxel-Based Methods.} 
Voxel-based methods have been an active research domain of 3D reconstruction over the past decades \cite{Seitz1999, Broadhurst2001, Bhotika2002, Pollard2007, Kolev2009, Liu2014, Savinov2015, Ulusoy2015}. 
Their core idea is to discretise the target volume of space in which the reconstruction is performed. 
With the renaissance of deep learning in the modern era \cite{Krizhevsky2012, Goodfellow2014, Szegedy2015, Jaderberg2015, He2016}, there have been multiple attempts to adapt voxel-based techniques to learning-based 3D reconstruction \cite{Choy2016, Riegler2017}. 
These methods have been criticised for a high training complexity due to expensive 3D convolutions 
and discretisation artefacts due to a low grid resolution. 
In contrast to previous works, we use a voxel-based proxy to regress displacements instead of deformed shapes. 
Note that in many related tasks, deformations are parametrised by lower-resolution data structures such as deformation graphs \cite{Sumner2007, NewcombeFS15, Xu2018}. 
To alleviate discretisation artefacts and enable superresolution of displacements, 
we apply point projection and trilinear interpolation. 

\paragraph{Learning Deformation Models.} 
Recently, the first supervised learning methods trained for a deformation model were proposed for monocular non-rigid 3D reconstruction \cite{Golyanik2018, Pumarola2018, Shimada2019, Tretschk2019arXiv}. 
Their main idea is to train a deep neural network (DNN) for feasible deformation modes from collections of deforming objects with known correspondences between non-rigid states. 
Implicitly, a single shape at rest (a thin surface) is assumed which is deformed upon 2D observations. 
Next, several works include a free-form deformation component for monocular rigid 3D reconstruction with an object-class template \cite{Kurenkov2018, Jack2018}. 
Hanocka \textit{et al.}~\cite{Hanocka2018} align meshes in a voxelised representation with an unsupervised learning approach. 
They learn a shape-aware deformation prior from shape datasets and can handle incomplete data.

Our method is inspired by the concept of a learned deformation model. 
In NRPSR, both the reference and template can differ significantly from scenario to scenario, and we cannot assume a single template for all alignment problems. 
To account for different scenarios and inputs, we introduce a proxy voxel grid which abstracts away from the point cloud representation. %
We learn a deformation model for displacements instead of a space of feasible deformation modes for shapes. 
Thus, we are able to use the same data modality for training as in \cite{Golyanik2018, Shimada2019, Tretschk2019arXiv} and generalise to arbitrary point clouds for non-rigid alignment. 
Wang \textit{et al.}~\cite{3dphysnetNIPS} solve a related problem on a voxel grid: predicting object deformations under applied physical forces. 
Their network is trained in an adversarial manner with ground truth deformations conditioned upon the elastic properties of the material and applied forces. 
In contrast to 3D-PhysNet \cite{3dphysnetNIPS}, we learn displacement fields on a voxel grid which is a more explicit representation of deformations intrinsic to NRPSR.

\section{The Proposed Approach}\label{sec:approach} 

We propose a neural network-based method for NRPSR that takes a template and a reference point set and returns a displacement vector for each template point, see Fig.~\ref{fig:architecture} for an overview. 
As described in Sec.~\ref{sec:related_work}, existing methods show lower accuracy in the presence of large deformations between point sets. 
We expect that neural-network-based methods are able to deal with such challenging cases since they learn class-specific priors implicitly during training. 
In NRPSR, the numbers of points in the template and reference are generally different. 
This inconsistency of the input dimensionality is problematic because we need to fix the number of neurons before training. 
To resolve this issue, we convert the point sets into a regular voxel-grid representation at the beginning of the pipeline, which makes our approach invariant with respect to the number and order of input points. 
Furthermore, due to the nature of convolutional layers, we expect a network with 3D convolutions to be robust to noises and outliers. 
Even though handling 3D voxel data is computationally demanding, modern hardware supports sufficiently fine-grained voxel grids which our approach relies on. %

\begin{figure}[t!]
 \centering
  \includegraphics[width=1.0\linewidth]{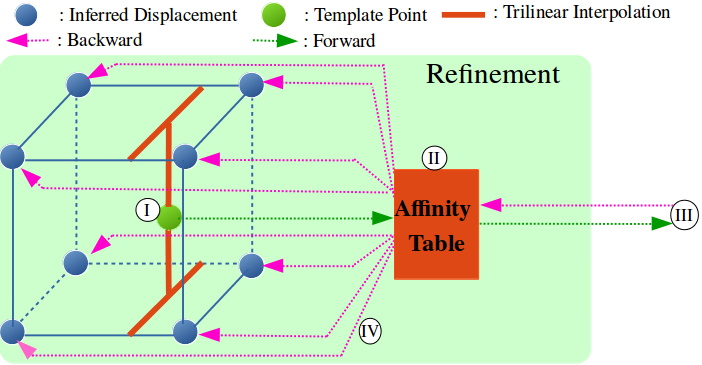} 
  \caption{Overview of the forward and backward pass of trilinear interpolation on a voxel grid. 
  Numbers \textcircled{\tiny{I}}, \textcircled{\tiny{II}}, \textcircled{\tiny{III}} and \textcircled{\tiny{IV}} indicate the sequence of steps performed in every training iteration, see Sec.~\ref{sec:architecture} for more details. 
  }
  \label{fig:grad_table} 
\end{figure}
\begin{table}
\center
\scalebox{0.64}{
\begin{tabular}{|c|c|c|c|c|c|}\hline
 \textbf{ID} & \textbf{Layer} & \textbf{Output Size} & \textbf{Kernel} & \textbf{Padding/Stride} & \textbf{Concatenation}\\\hline
 1  &     Input & 64x64x64x2 & - & - & - \\\hline
 2  &     3D Convolution & 64x64x64x8 & 7x7x7 & 3/1 & -\\\hline %
 3  &     LeakyReLU & 64x64x64x8 & - & -  & -\\\hline
 4  &     MaxPooling 3D & 32x32x32x8 & 2x2x2 & 0/2 & -\\\hline
 
 5  &     3D Convolution & 32x32x32x16 & 5x5x5 & 2/1 & -\\\hline %
 6  &     LeakyReLU & 32x32x32x16 & - & - & -\\\hline
 7  &     MaxPooling 3D & 16x16x16x16 & 2x2x2 & 0/2 & -\\\hline
 
 8  &     3D Convolution & 16x16x16x32 & 3x3x3 & 1/1 & -\\\hline %
 9  &     LeakyReLU & 16x16x16x32 & - & - & -\\\hline
 10 &     MaxPooling 3D & 8x8x8x32 & 2x2x2 & 0/2 & -\\\hline
 
 11 &     3D Convolution & 8x8x8x64 & 3x3x3 & 1/1 & -\\\hline %
 12 &     LeakyReLU & 8x8x8x64 & - & - & -\\\hline
     
 13 &     3D Deconvolution & 16x16x16x64 & 2x2x2 & 0/2 & 12 \& 10 \\\hline %
 14 &     3D Deconvolution & 16x16x16x64 & 3x3x3 & 1/1 & -\\\hline 
 15 &     LeakyReLU & 16x16x16x64 & - & - & -\\\hline
     
 16 &     3D Deconvolution & 32x32x32x32 & 2x2x2 & 0/2 & 15 \& 7 \\\hline %
 17 &     3D Deconvolution & 32x32x32x32 & 5x5x5 & 2/1 & -\\\hline 
 18 &     LeakyReLU & 32x32x32x32 & - & - & - \\\hline
     
 19 &     3D Deconvolution & 64x64x64x16 & 2x2x2 & 0/2 & 18 \& 4 \\\hline %
 20 &     3D Deconvolution & 64x64x64x16 & 7x7x7 & 3/1 & -\\\hline 
 21 &     LeakyReLU & 64x64x64x16 & - & - & -\\\hline
 22 &     3D Deconvolution & 64x64x64x3 & 3x3x3 & 1/1 & -\\\hline %
 
 \end{tabular}
 }
 \caption{U-Net-style architecture of \hbox{DispVoxNet}. The concatenation column contains the layer IDs whose outputs are concatenated and used as an input to the current layer. 
 We use a negative slope for LeakyReLU of $0.01$.}
 \label{tab:architecture_detail}
\end{table}

\paragraph{Notations and Assumptions.} 
The inputs of our algorithm are two point sets: the reference $\bfX = (\bfx_1, \hdots, \bfx_N )^\mathsf{T} \in \mathbb{R}^{N \times D}$ and the template $\bfY = (\bfy_1, \hdots, \bfy_M)^\mathsf{T} \in \mathbb{R}^{M \times D}$ which has to be non-rigidly matched to $\bfX$. 
$N$ and $M$ are the cardinalities of $\bfX$ and $\bfY$ respectively, and $D$ denotes the point set dimensionality. 
We assume the general case when $M \neq N$ and 
$D = 3$ in all experiments, although our method is directly applicable to $D = 2$ and generalisable to $D > 3$ if training data is available and a voxel grid is feasible in this dimension. 
Our objective is to find the displacement function (a vector field) $v: \mathbb{R}^{M \times 3} \times \mathbb{R}^{N \times 3} \rightarrow \mathbb{R}^{M \times 3}$ so that  $\bfY + v(\bfY,\bfX)$ matches $\bfX$ as close as possible.

There is no universal criterion for optimal matching and it varies from scenario to scenario. 
We demand 1) that $v$ results in realistic class-specific object deformations so that the global alignment is recovered along with fine local deformations, and 2) that the template deformation preserves the point topology as far as possible. 
The first requirement remains very challenging for current general NRPSR methods. 
Either the shapes are globally matched while fine details are disregarded or the main deformation component is neglected which can lead to distorted registrations. 
The problem becomes even more ill-posed due to noise in the point sets. 
Some methods apply multiscale matching or parameter adjustment schemes \cite{Granger2002, Golyanik2016}. %
Even though a relaxation of the global topology-preserving constraint can lead to a finer local alignment, there is an increased chance of arriving at a local minimum and fitting to noise.

Let $\bfV_{\bfX}$ and $\bfV_{\bfY}$ be the voxel grids, \textit{i.e.,} the voxel-based proxies on which \hbox{DispVoxNets} regress
deformations.
Without loss of generality, we assume that $\bfV_{\bfX}$ and $\bfV_{\bfY}$ are cubic and both have equal dimensions $Q=64$. We propose to learn $v$ as described next. 

\subsection{Architecture}\label{sec:architecture} 

Our method is composed of \emph{displacement estimation} and \emph{refinement} stages. 
Each stage contains a DispVoxNet, \textit{i.e.,} a 3D-convolutational neural network based on a U-Net architecture %
~\cite{ronneberger2015u} %
, which we also denote by $\mathbf{D_{vn}}$. 
See Figs.~\ref{fig:architecture}--\ref{fig:grad_table} and  Table~\ref{tab:architecture_detail} for details on the network architecture. 

\paragraph{Displacement Estimation (DE) Stage.} 
We first discretise both $\bfY$ and $\bfX$ on $\bfV_{\bfY}$ and $\bfV_{\bfX}$, respectively (\emph{P2V} in Fig.~\ref{fig:architecture}). During the conversion, the point-to-voxel correspondences are stored in an affinity table. 
As voxel grids sample the space uniformly, each point of $\bfX$ and $\bfY$ is mapped to one of the $Q^3$ voxels in $\bfV_{\bfX}$ and $\bfV_{\bfY}$, respectively. %
In $\bfV_{\bfX}$ and $\bfV_{\bfY}$, we represent $\bfX$ and $\bfY$ as binary voxel occupancy indicators, \textit{i.e.,} if at least one point falls into a voxel, the voxel's occupancy is set to $1$; otherwise, it equals to $0$. 
Starting from $\bfV_{\bfX}$ and $\bfV_{\bfY}$, \hbox{DispVoxNet} regresses 
per-voxel displacements of the dimension $Q^3 \times 3$. 
During training, we penalise the discrepancy between the inferred voxel displacements and the ground truth displacements $\mathbf{Z}$ using a mean squared error normalised by the number of voxels. 
$\mathbf{Z}$ is obtained by converting the ground truth point correspondences to the voxel-based representation of dimensions $Q^3 \times 3$ compatible with our architecture. 
The \textit{point displacement} loss is given by: 
\begin{equation}
\mathbf{\mathcal{L}_{Disp.}}(\mathbf{Z},\bfV_{\bfY},\bfV_{\bfX})=\frac{1}{Q^{3}}\left\Vert\mathbf{Z} - \mathbf{D_{vn}(\bfV_{\bfY},\bfV_{\bfX})}\right\Vert_{2}^{2}.
\end{equation}
Using the affinity table between $\bfY$ and $\bfV_{\bfY}$, we determine each point's displacement by applying trilinear interpolation on the eight nearest displacements in the voxel grid (\emph{V2P} in Fig.~\ref{fig:architecture} and see supplement). 
After adding the displacements to $\bfY$, we observe that the resulting output after a single DispVoxNet bears some roughness. The refinement stage described in the following %
alleviates this problem. 
 \begin{figure}[t!]
 \centering
  \includegraphics[width=1.0\linewidth]{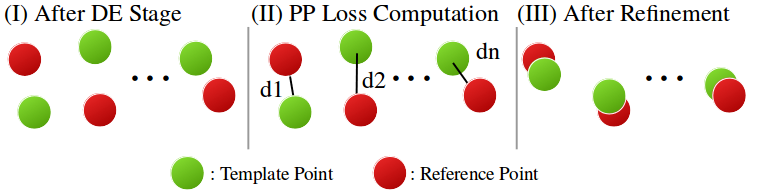} 
  \caption{(I) 
  The displacement estimation stage regresses 
  rough displacements between $\bfY$ and $\bfX$. 
  (II) 
  For all $\bfy'$ in $\bfY + v(\bfY,\bfX)$, we find the distance to the closest point $\bfx_{\bfy'}$ among all points in $\bfX$. %
  (III) 
  At test time, the refined displacements yield a smoothed result $\bfY + v(\bfY,\bfX)$.
  }
  \label{fig:PPLoss} 
\end{figure} 

\paragraph{Refinement Stage.}

Since the %
DE stage accounts for global deformations but misses some fine details, the unresolved residual displacements at the refinement stage are small. 
Recall that \hbox{DispVoxNet} is exposed to scenarios with small displacements during the training of the DE stage, since our datasets also contain similar (but not 
equal) states. 
Thus, assuming small displacements, we design a refinement stage as a combination of a pre-trained \hbox{DispVoxNet} and an additional unsupervised loss. 
Eventually, the refinement stage resolves the remaining small displacements and smooths the displacement field.  
To summarise, at the beginning of the refinement stage, the already deformed template point set is converted into the voxel representation $\bfV_{\bfY}^{*}$. 
From $\bfV_{\bfY}^{*}$ and $\bfV_{\bfX}$, a pre-trained \hbox{DispVoxNet} %
learns to regress refined per-voxel displacements.

To apply the inferred voxel displacements to a template point at the end of the refinement stage (see Fig.~\ref{fig:grad_table}), (I)\footnote{(I), (II), (III) and (IV) refer to the steps in Fig.~\ref{fig:grad_table}} we compute the trilinear interpolation of the eight nearest displacements of $\bfy_i$, $i \in \{1, \hdots, M\}$, 
and 
calculate a weighted consensus displacement for $\bfy_i$. %
(II) The weights and indices of the eight nearest voxels are saved in an affinity table. 
To further increase the accuracy, 
(III) we introduce the unsupervised, differentiable \textit{point projection} (PP) loss between the final output $\bfY + v(\bfY,\bfX)$ and $\bfX$. 
The PP loss penalises the Euclidean distances between a point $\bfy'$ in $\bfY + v(\bfY,\bfX)$ and its closest point  $\bfx_{\bfy'}$ in $\bfX$: 
\begin{equation}
\mathbf{\mathcal{L}_{PP}}(\bfY + v(\bfY,\bfX),\bfX)=\frac{1}{M}\sum_{i=1}^{M}\|\bfy'_{i} - \bfx_{\bfy'}\|_{2}.
\end{equation}
We employ a \textit{k}-d tree to determine $\bfx_{\bfy'}$ for all $\bfy'$ in $\bfY + v(\bfY,\bfX)$, see Fig.~\ref{fig:PPLoss} for a schematic visualisation.

Since the training is performed through backpropagation, we need to ensure the differentiability of all network stages. 
Our approach contains conversions from voxel to point-set representations and vice versa that are not fully differentiable. %
Thanks to the affinity table, we know the correspondences between points and voxels at the refinement stage. 
Therefore, (IV) gradients back-propagated from the PP loss can be distributed into the corresponding voxels in the voxel grid as shown in Fig.~\ref{fig:grad_table}. 
As eight displacements contribute to the displacement of a point %
due to trilinear interpolation in the forward pass, the gradient of the point is back-propagated to the eight nearest voxels in the voxel grid according to the trilinear weights from the forward pass. %

Two consecutive \hbox{DispVoxNets} in the DE and refinement stages implement a hierarchy with two granularity levels for the regression of point displacements. Combining more stages does not significantly improve the result, \textit{i.e.,} we find two \hbox{DispVoxNets} necessary and sufficient.

\subsection{Training Details}\label{sec:training_detail}

We use Adam \cite{kingma2014adam} optimiser with a learning rate of \hbox{$3\cdot 10 ^ {-4}$}. 
As the number of points varies between the training pairs, we set the batch size to 1. 
We train the stages in two consecutive phases, starting with the  
DE stage using the displacement loss until convergence. 
This allows the network to regress rough displacements between $\bfY$ and $\bfX$. 
Then, another instance of \hbox{DispVoxNet} in the refinement stage is trained using only the PP loss. 
We initialise it with the weights from \hbox{DispVoxNet} of the DE stage. 
Since the PP loss considers only one nearest neighbour, 
we need to ensure that each output point from the DE stage is already close to its corresponding point in $\bfX$. 
Thus, we freeze the weights of \hbox{DispVoxNet} in the DE stage when training the refinement stage. See %
our supplement for training statistics. 

To enhance robustness of the network to noise and missing points, $0-30\%$ of the points are removed at random from $\bfY$ and $\bfX$, and uniform noise is added to both point sets. The number of added points ranges from 0\% to 100\% of the number of points in the respective point set. 
The amount of noise per sample is determined randomly. 
When computing the PP loss, added noise is not considered.

\section{Experiments}\label{sec:experiments} 

\begin{table}
\center
\scalebox{0.73}{
\begin{tabular}{|cc|cccc|}\hline
				                                            &               & \textbf{Ours}            & \textbf{NR-ICP} \cite{Chui2003}  & \textbf{CPD} \cite{Myronenko2010}	           & \textbf{GMR} \cite{Jian2005} \\  \hline
      \multirow{2}{*}{\textbf{thin plate}\cite{Golyanik2018}} 		& $e$		    & 0.0103		  & 0.0402			& \textbf{0.0083} / 0.0192   & 0.2189	\\ 	\cline{2-6}
                		                                    & $\sigma$		& \textbf{0.0059} & 0.0273			& 0.0102 / 0.0083            & 1.0121	\\ \hline
      \multirow{2}{*}{\textbf{FLAME}\cite{FLAME:SiggraphAsia2017}} 	& $e$		    & 0.0063		  & 0.0588			& \textbf{0.0043} / 0.0094   & 0.0056	\\ 		\cline{2-6}
                	            	                        & $\sigma$		& 0.0009		  & 0.0454			& 0.0008 / \textbf{0.0005}   & 0.0007    \\ \hline
      \multirow{2}{*}{\textbf{DFAUST}\cite{dfaust2017}} 	    & $e$		    & \textbf{0.0166} & 0.0585	        & 0.0683 / 0.0721            & 0.2357	\\ 		\cline{2-6}
            		                                        & $\sigma$		& \textbf{0.0020} & 0.0215			& 0.0314 / 0.0258            & 0.8944 	\\ \hline
      \multirow{2}{*}{\textbf{cloth}\cite{bednarik2018learning}} 	& $e$		    & \textbf{0.0080} & 0.0225			& 0.0149 / 0.0138            & 0.2189	\\ 		\cline{2-6}
                	                 	                    & $\sigma$		& \textbf{0.0021} & 0.0075			& 0.0066 / 0.0033            & 1.0121	\\ \hline
 \end{tabular}
 }
 \caption{\label{tab:base_comp} Comparison of registration errors for all tested methods. 
 For CPD, we also report results with FGT (right-hand values). 
 } 
\end{table}

\begin{table} 
\center 
 \scalebox{0.75}{ 
 \begin{tabular}{|c|c|c|c|}\hline
				           & \textbf{DE} & \textbf{DE + Ref. (nearest voxel)} & \textbf{Full: DE + Ref. (trilinear)}\\ \hline
      $e$		    & 0.0100  	    & 0.0088			& \textbf{0.0069} \\ \hline
      $\sigma$		& 0.0021		& 0.0075			& \textbf{0.0016}      \\ \hline
\end{tabular}
}
\caption{\label{tab:ablation} Ablation study highlighting the importance of the refinement stage and trilinear interpolation compared to a nearest voxel lookup.} 
\end{table}

Our method is implemented in PyTorch \cite{paszke2017automatic}. 
The evaluation system contains two Intel(R) Xeon(R) E5-2687W v3 running at 3.10GHz 
and a NVIDIA GTX 1080Ti GPU. %
We compare \hbox{DispVoxNets} with four methods with publicly available code, \textit{i.e.,}~point-to-point non-rigid ICP (NR-ICP) \cite{Chui2003}, GMR \cite{Jian2005}, CPD \cite{Myronenko2010} and CPD with Fast Gaussian Transform (FGT) \cite{Myronenko2010}. 
FGT is a technique for the fast evaluation of Gaussians at multiple points \cite{Greengard1991}.

\subsection{Datasets}\label{sec:dataset} 
In total, we evaluate on four different datasets which represent various types of common 3D deformable objects, \textit{i.e.,} %
\textit{thin plate} \cite{Golyanik2018}, \textit{FLAME} \cite{FLAME:SiggraphAsia2017}, \textit{Dynamic FAUST (DFAUST)} \cite{dfaust2017} and \textit{cloth} \cite{bednarik2018learning}. 
\emph{Thin plate} contains $4648$ states of a synthetic isometric surface. 
\emph{FLAME} consists of a variety of captured human facial expressions ($10k$ meshes in total). 
\emph{DFAUST} is a scanned mesh dataset of human subjects in various poses %
($7948$ meshes in total). 
Lastly, the \emph{cloth} dataset %
contains $6238$ captured states of a deformable sheet. 
Except for \textit{FLAME}, the datasets are sequential and contain large non-linear deformations. 
Also, the deformation complexity in \textit{FLAME} is lower than in the other datasets, \textit{i.e.,} the deformations are mostly concentrated around the mouth area of the face scans. 

We split the datasets into training and test subsets by considering blocks of a hundred point clouds. %
The first eighty samples from every block comprise the training set and the remaining twenty are included in the test set. 
For \textit{FLAME}, we pick $20\%$ of samples at random for testing and use the remaining ones for training. 
As all datasets have consistent topology, we directly obtain ground truth correspondences which are necessary for training and error evaluation. 
We evaluate the registration accuracy of our method on clean samples (see Sec.~\ref{exp:noiseless_data}) as well as in settings with uniform noise and clustered outliers (added sphere and removed chunk, see Sec.~\ref{sec:outlier_chunkRem}), since point cloud data captured by real sensors often contains noise and can be incomplete. 
In total, thirty template-reference pairs are randomly selected from each test dataset. 
We use the same pairs in all experiments. 
For the selected pairs, we report the average root-mean-square error (RMSE) between the references and aligned templates 
and standard deviation of RMSE, denoted by $e$ and $\sigma$ respectively: 
$e = \frac{1}{M}\sum_{i=1}^{M}\frac{\left \| \bfy_{i}-\bfx_{i} \right \|_{2}}{\sqrt{D}},$
with the template points $\bfy_{i}$ and corresponding reference points $\bfx_{i}$.

\begin{figure*}[t!]
 \centering
  \includegraphics[width=1.0\linewidth]{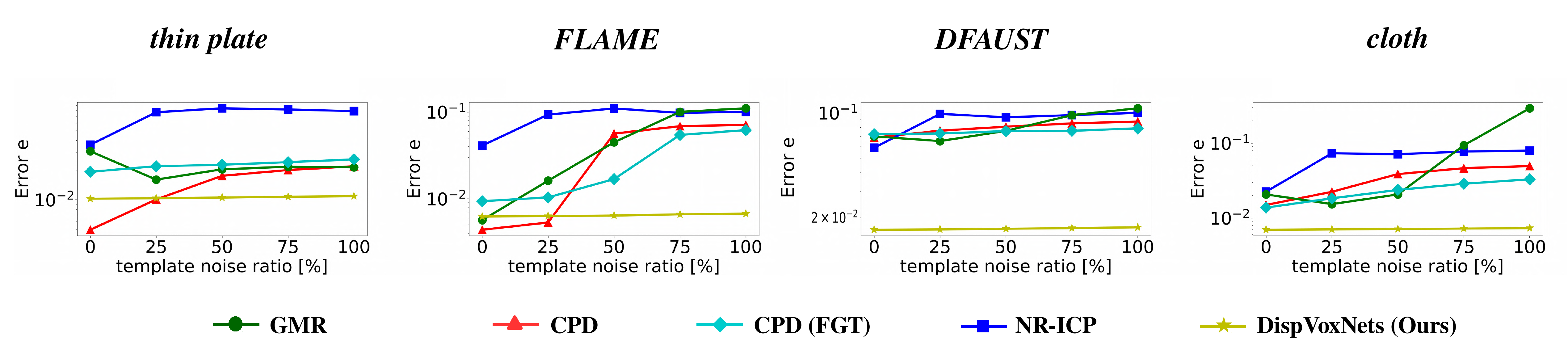} 
  \caption{Comparison of registration errors in the scenario with uniform noise. 
  $p\%$ is the ratio between the number of noise points added to the template and the number of points in the unperturbed template. 
  $e$ statistics of our approach is barely affected by the noise level. 
  See 
  the supplement 
  for more results. 
  }
  \label{fig:uni_noise} 
\end{figure*}

\begin{table*}[t]
        \begin{minipage}{0.5\textwidth}
            \centering
            \scalebox{0.66}{
            \begin{tabular}{|ccc|cccc|}\hline
				                                            &                               &               & \textbf{Ours}              & \textbf{NR-ICP} \cite{Chui2003}    & \textbf{CPD} \cite{Myronenko2010}	            & \textbf{GMR} \cite{Jian2005} \\ \hline
      \multirow{4}{*}{\textbf{\textit{thin plate}}\cite{Golyanik2018}}        & \multirow{2}{*}{ref.}	& $e$		    & \textbf{0.0151} 	& 0.0349		    & 0.1267 / 0.1136		    & 0.6332	\\ 	\cline{3-7}
                                    		                &                               & $\sigma$		& \textbf{0.0117}	& 0.0302		    & 0.0224 / 0.0211		    & 1.5749	\\ \cline{2-7}
                                                            & \multirow{2}{*}{temp.}	    & $e$		    & \textbf{0.0150}	& 0.0509		    & 0.0304 / 0.0636 		    & 0.0528	\\ 	\cline{3-7}
                                    		                &                               & $\sigma$		& \textbf{0.0106}	& 0.0406		    & 0.0200 / 0.0149		    & 0.0300	\\ \hline
      \multirow{4}{*}{\textbf{\textit{FLAME}}\cite{FLAME:SiggraphAsia2017}}   & \multirow{2}{*}{ref.}	& $e$		    & 0.0098	        & \textbf{0.0039}	& 0.0492 / 0.0617 		    & 0.0577	\\ 	\cline{3-7}
                                            		        &                               & $\sigma$		& 0.0034 	        & \textbf{0.0007}	& 0.0301 / 0.0218		    & 0.0205	\\ \cline{2-7}
                                                            & \multirow{2}{*}{temp.}	    & $e$		    & 0.0073	        & 0.0566		    & \textbf{0.0072} / 0.0246 	& 0.0309	\\ 	\cline{3-7}
                                            		        &                               & $\sigma$		& \textbf{0.0015}	& 0.0334		    & 0.0070 / 0.0142	        & 0.0117	\\ \hline
      \multirow{4}{*}{\textbf{\textit{DFAUST}}\cite{dfaust2017}}         & \multirow{2}{*}{ref.}	& $e$		    & \textbf{0.0308}	& 0.0605		    & 0.1127 / 0.1151           & 0.9730  	\\ 	\cline{3-7}
                                    		                &                               & $\sigma$		& \textbf{0.0111}	& 0.0226		    & 0.0308 / 0.0295		    & 2.2267	\\ \cline{2-7}
                                                            & \multirow{2}{*}{temp.}	    & $e$		    & \textbf{0.0190}	& 0.0669     	    & 0.0791 / 0.0775		    & 0.0845	\\ 	\cline{3-7}
                                    		                &                               & $\sigma$		& \textbf{0.0036}	& 0.0187		    & 0.0304 /	0.0220	        & 0.0295    \\ \hline
      \multirow{4}{*}{\textbf{\textit{cloth}}\cite{bednarik2018learning}}   & \multirow{2}{*}{ref.}	& $e$		    & \textbf{0.0213}	& 0.0248		    & 0.1081 / 	0.1096	        & 0.1098	\\ 	\cline{3-7}
                                            		        &                               & $\sigma$		& \textbf{0.0091}	& 0.0095		    & 0.0235 /	0.0223          & 0.0234	\\ \cline{2-7}
                                                            & \multirow{2}{*}{temp.}	    & $e$		    & 0.0649	        & \textbf{0.0296}	& 0.0408 /	0.0522	        & 0.0476	\\ 	\cline{3-7}
                                            		        &                               & $\sigma$		& 0.0395	        & \textbf{0.0081}	& 0.0115 /	0.0114	        & 0.0223	\\ \hline
 \end{tabular}
 }
 \captionsetup{width=0.95\textwidth}
            \caption{\label{tab:outlier} Registration errors for the case with clustered outliers. 
 For CPD, we also report results in the mode with FGT (right-hand values). 
 \textit{``ref."} and \textit{``temp."} denote whether outliers are added to $\bfX$ or $\bfY$, respectively. 
 }
        \end{minipage} 
        \hfill
        \begin{minipage}{0.5\textwidth}
            \centering
            \scalebox{0.66}{
            \begin{tabular}{|ccc|cccc|}\hline
			                                                &                               &               & \textbf{Ours}              & \textbf{NR-ICP} \cite{Chui2003}& \textbf{CPD} \cite{Myronenko2010}	            & \textbf{GMR} \cite{Jian2005} \\ \hline
      \multirow{4}{*}{\textbf{\textit{thin plate}}\cite{Golyanik2018}}        & \multirow{2}{*}{ref.}	& $e$		    & \textbf{0.0107}	& 0.0668		& 0.0218 / 0.0386		    & 0.4415	\\ 	\cline{3-7}
                                    		                &                               & $\sigma$		& \textbf{0.0061}	& 0.0352		& 0.0148 / 0.0067           & 1.4632	\\ \cline{2-7}
                                                            & \multirow{2}{*}{temp.}	    & $e$		    & \textbf{0.0108}	& 0.0334		& 0.0479 / 0.0471		    & 0.4287	\\ 	\cline{3-7}
                                    		                &                               & $\sigma$		& 0.0062            & 0.0281		& 0.0101 / \textbf{0.0038}  & 1.3832	\\ \hline
      \multirow{4}{*}{\textbf{\textit{FLAME}}\cite{FLAME:SiggraphAsia2017}}   & \multirow{2}{*}{ref.}	& $e$		    & 0.0084	        & 0.0519		& \textbf{0.0046} / 0.0140  & 0.0193	\\ 	\cline{3-7}
                                            		        &                               & $\sigma$		& 0.0010	        & 0.0451		& 0.0009 / \textbf{0.0006}	& 0.0008	\\ \cline{2-7}
                                                            & \multirow{2}{*}{temp.}	    & $e$		    & 0.0088	        & 0.0215		& \textbf{0.0076} / 0.0201	& 0.0274	\\ 	\cline{3-7}
                                            		        &                               & $\sigma$		& \textbf{0.0010}	& 0.0219		& \textbf{0.0010} / 0.0016	& 0.0019	\\ \hline
      \multirow{4}{*}{\textbf{\textit{DFAUST}}\cite{dfaust2017}}         & \multirow{2}{*}{ref.}	& $e$		    & \textbf{0.0167}	& 0.0463		& 0.0562 / 0.0636	        & 0.0714    \\ 	\cline{3-7}
                                        		            &                               & $\sigma$		& \textbf{0.0029}	& 0.0195		& 0.0308 / 0.0216	        & 0.0282    \\ \cline{2-7}
                                                            & \multirow{2}{*}{temp.}	    & $e$		    & \textbf{0.0169}	& 0.0426    	& 0.0672 / 0.0710           & 0.0737    \\ 	\cline{3-7}
                                        		            &                               & $\sigma$		& \textbf{0.0033}	& 0.0194		& 0.0291 / 0.0229           & 0.0243    \\ \hline
      \multirow{4}{*}{\textbf{\textit{cloth}}\cite{bednarik2018learning}}   & \multirow{2}{*}{ref.}	& $e$		    & \textbf{0.0090}	& 0.0455		& 0.0248 / 0.0315 	        & 0.0288	\\ 	\cline{3-7}
                                            		        &                               & $\sigma$		& \textbf{0.0018}	& 0.0061		& 0.0056 / 0.0027	        & 0.0087	\\ \cline{2-7}
                                                            & \multirow{2}{*}{temp.}	    & $e$		    & \textbf{0.0132}	& 0.0208		& 0.0486 / 0.0347		    & 0.0397	\\ 	\cline{3-7}
                                            		        &                               & $\sigma$		& 0.0019	        & 0.0087		& 0.0077 / \textbf{0.0014}		    & 0.0092	\\ \hline
 \end{tabular}
 }
 \captionsetup{width=0.95\textwidth}
            \caption{\label{tab:chunk_removal} Registration errors for the case with missing parts. 
 For CPD, we also report results in the mode with FGT (right-hand values).
 \textit{``ref."} and \textit{``temp."} denote whether parts are removed from $\bfX$ or $\bfY$, respectively. 
 }
        \end{minipage}
    \end{table*}

\subsection{Noiseless Data}\label{exp:noiseless_data}  %
We first evaluate the registration accuracy of our method and several baselines on noiseless data. Table~\ref{tab:base_comp} and Fig.~\ref{fig:teaser} summarise the results. %
Our approach significantly outperforms other methods on the \textit{DFAUST} and \textit{cloth} datasets which contain articulated motion and large non-linear deformations between the template and reference. 
On \textit{thin plate}, \hbox{DispVoxNets} perform on par with CPD (CPD with FGT) and show a lower $e$ in three cases out of four. 
On \textit{FLAME}, which contains localised and small deformations, our approach achieves $e$ of the same order of magnitude as CPD and GMR.  
CPD, GMR and \hbox{DispVoxNets} outperform NR-ICP in all cases. 
The experiment confirms the advantage of our approach in aligning point sets with large global non-linear deformations (additionally, see %
the supplement).

\paragraph{Ablation Study.} 
We conduct an ablation study on the \textit{cloth} dataset to test the influence of each component of our architecture. 
The tested cases include 1) only DE stage, 2) DE and refinement stages with nearest voxel lookup (the na\"ive alternative to trilinear interpolation) 
and 3) our entire architecture with DE and refinement stages, plus trilinear interpolation. 
Quantitative results  are shown in Table~\ref{tab:ablation}. 
The full model with trilinear interpolation reduces the error by more than $30\%$ over the DE only setting.

\subsection{Deteriorated Data}\label{sec:outlier_chunkRem} 

The experiments with deteriorated data follow the evaluation protocol of  Sec.~\ref{exp:noiseless_data}. 
We introduce clustered outliers and uniform noise to the data.

\paragraph{Structured  Outliers.} 
We evaluate the robustness of our method to added structured outliers and missing data. 
We either add a sphere-like object to the inputs or arbitrarily remove a chunk from one of the point sets. 
As summarised in Tables \ref{tab:outlier}--\ref{tab:chunk_removal}, our approach shows the highest accuracy among all methods on \textit{thin plate}, \textit{DFAUST} and \textit{cloth}, even though \hbox{DispVoxNets} were not trained with clustered outliers and have not been exposed to sphere-like structures. 

\begin{figure*}[t!] 
 \centering 
  \includegraphics[width=1.0\linewidth]{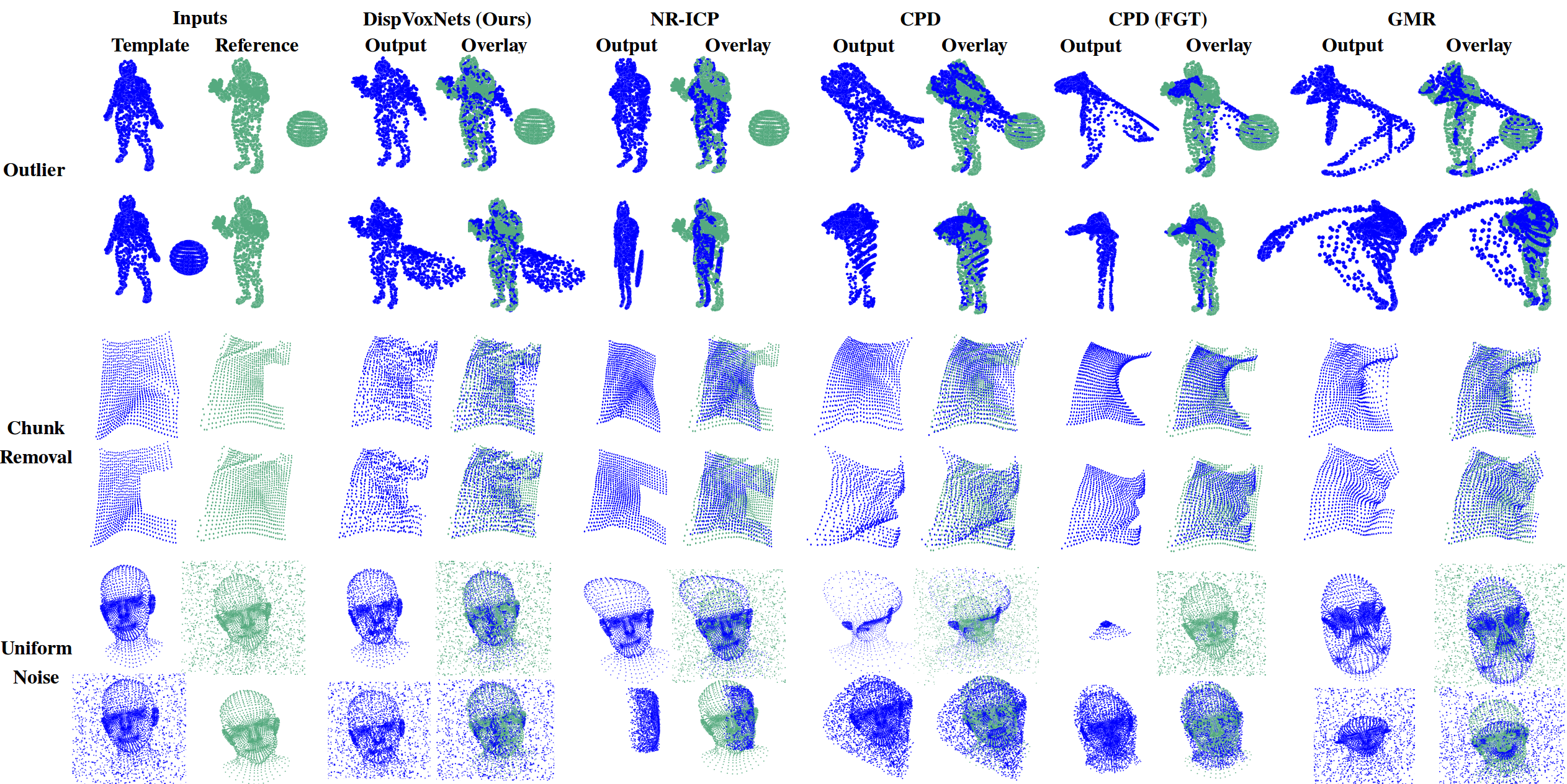} 
  \caption{Qualitative comparison of our approach to other methods in challenging scenarios with added clustered outliers (sphere added to either template or reference; the first and second rows), removed parts (the third and fourth rows) and $50\%$ of added uniform noise (the fifth and sixth rows). 
  } 
  \label{fig:outlier_comp} 
\end{figure*}

\noindent Tables \ref{tab:outlier}--\ref{tab:chunk_removal} report results for both the cases with modified references and templates, and Fig.~\ref{fig:outlier_comp} compares results qualitatively. 
\hbox{DispVoxNets} are less influenced by the outliers and produce visually accurate alignments of regions with correspondences in both point sets. 
CPD, GMR and NR-ICP suffer from various effects, \textit{i.e.,} their alignments are severely influenced by outliers and the target regions are corrupted in many cases. 
We hypothesise that convolutional layers in DispVoxNet learn to extract informative features from the input points set and ignore noise. %
Furthermore, the network learns a class-specific deformation model which further enhances the robustness to outliers. 

\paragraph{Uniform Noises.} %
Next, we augment templates with uniform noise and repeat the experiment. 
Fig.~\ref{fig:uni_noise} reports metrics for different amounts of added noise. %
Note that CPD, GMR and NR-ICP fail multiple times, and we define the success criterion in this experiment
as $e < (4\times\text{median})$ followed by $e < 4.0$.
\hbox{DispVoxNets} show stable accuracy across different noise ratios and datasets, while the error of other approaches increases significantly (up to $100$ times) with an increasing amount of noise. 
Only our approach is agnostic to large amount of noise, despite CPD explicitly modeling a uniform noise component. 
For a qualitative comparison, see the fifth and sixth rows in Fig.~\ref{fig:outlier_comp}  
as well as the supplement.

\subsection{Runtime Analysis} 
\begin{figure}[t!]
 \centering
  \includegraphics[width=0.9\linewidth]{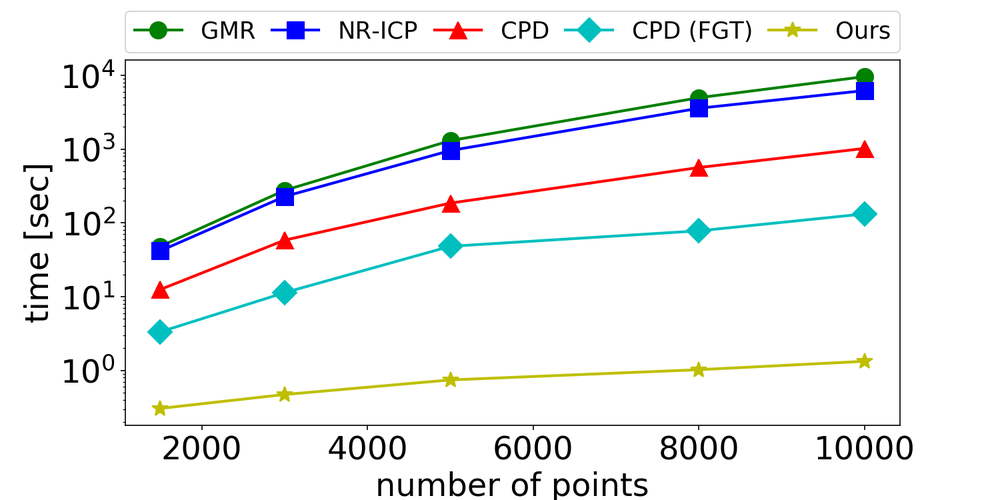} 
  \caption{Runtime comparison for different number of points. The horizontal axis shows the number of points in the template and reference point sets. %
  The vertical axis shows the $\log_{10}$-scaled inference time in seconds.} 
  \label{fig:runtime} 
\end{figure}

We prepare five point set pairs out of \textit{DFAUST} dataset where the number of points varies from $1.5k$ to $10k$. %
The runtime plots for different number of points are shown in Fig.~\ref{fig:runtime}. 
The numbers of points in both point sets are kept equal in this experiment. 
For $10k$ points, GMR, NR-ICP, CPD and CPD (FGT) take about 2 hours, 1.5 hours, 15 minutes, and 2 minutes per registration, respectively. 
Our approach requires only 1.5 seconds, which suggests its potential for applications with real-time constraints. 

\subsection{Real Scans} 
\begin{figure}%
 \centering
  \includegraphics[width=1.0\linewidth]{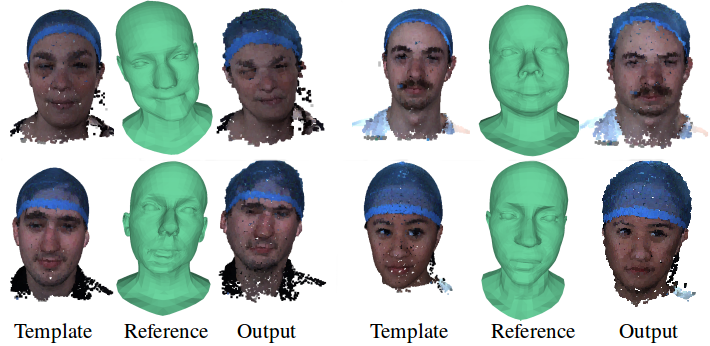} 
  \caption{
  Aligning a real scan from \cite{Dai_2017_ICCV} to a \textit{FLAME} \cite{FLAME:SiggraphAsia2017} reference. 
  The references are shown as meshes for visualisation purposes.  %
  }
  \label{fig:real_face} 
\end{figure}

In this section, we demonstrate the generalisability of \hbox{DispVoxNets} to real-world scans. 
We test on 3D head point sets from the data collection of Dai \textit{et al.}~\cite{Dai_2017_ICCV}. 
Since facial expression do not vary much in it, 
we use a reference from \textit{FLAME} \cite{FLAME:SiggraphAsia2017} 
and a template from \cite{Dai_2017_ICCV}. 
Registration results can be seen in Fig.~\ref{fig:real_face}. 
Although some distortion in the output shape is recognisable, \hbox{DispVoxNets} transfer the reference facial expression to the template. 
Even though the network has only seen \textit{FLAME} at training time, it is able to align two point sets of different cardinalities and origins.

\section{Conclusions}\label{sec:conclusions} 
We propose a DNN architecture with \hbox{DispVoxNets} --- the first NRPSR approach with a supervised learning proxy --- which regresses 3D displacements on regularly sampled voxel grids. 
Thanks to two consecutive \hbox{DispVoxNets} with trilinear interpolation and point projection, 
our approach outperforms other NRPSR methods by a large margin in scenarios 
with large deformations, articulations, clustered outliers and large amount of noise, while not relying on engineered class-specific priors. 
The runtime of \hbox{DispVoxNets} is around one second whereas other methods can take a few hours per registration, which suggests that our approach can be used in interactive applications. %
We show a high degree of generalisability of our architecture to real scans. 

We believe that the direction revealed in this paper has a high potential for further investigation. 
Even though \hbox{DispVoxNets} represent a promising step towards accurate non-rigid point set alignment, its accuracy is limited by the resolution of the voxel grid and composition of training datasets. 
In future work, our method could be extended for operation on non-uniform voxel grids, and other types of losses could be investigated. 
Embedding of further alignment cues such as point normals, curvature, colours as well as sparse prior matches is a logical next step. 
We also expect to see extensions of \hbox{DispVoxNets} for point sets with small overlaps and adaptations of the proposed architecture for learning-based depth map fusion systems.

{\small
\bibliographystyle{ieee}
\bibliography{egbib}
}

\normalsize 

\clearpage
\renewcommand{\thetable}{\Roman{table}}   
\renewcommand\thefigure{\Roman{figure}}  
\setcounter{figure}{0}   
\setcounter{table}{0} 
\begin{alphasection}
\section{Appendix} %
In this supplement, we provide details on the interpolation of the coarse displacement field  (Sec.~\ref{sec:trilinear}) and 
report training statistics (Sec.~\ref{sec:train_time}).
We show more qualitative comparisons (Sec.~\ref{sec:qual_comp}) as well as graphs for further cases with uniform noise (Sec.~\ref{sec:uni_noise}).

\subsection{Interpolation of the 3D Displacement Field}\label{sec:trilinear}

Due to the limited resolution of the voxel grid, we apply trilinear interpolation to obtain displacements for every template point at sub-voxel precision. 
Note that in DE stage, interpolation is applied only in the forward pass. 
In the refinement stage, it is applied in the forward pass, and the computed trilinear weights are used during backpropagation to weight the gradients. %

Suppose $\vec{\bfD}\colon \mathbb{Z}^3 \to \mathbb{R}^3$ 
is the initial regressed 3D displacement field on a regular lattice induced by the voxel grid. 
Suppose the template point of interest after the DE stage $\bfy_j^* = (x_j, y_j, z_j)$, $j \in \{1, \hdots, M\}$, falls into a neighbourhood cube between eight displacement values of $\vec{\bfD}$. 
We denote these boundary displacements compactly by $\bfd = \{ \bfd_{abc}\}$, $a, b, c \in \{0, 1\}$ on a unit cube\footnote{$\bfd_{abc}$ is a shorthand notation for the displacement at point $(x, y, z)$ in the local coordinate system, \textit{i.e.,} at $(0,0,0)$, $(0,0,1)$, $(0,1,0)$, \textit{etc.} } in a local coordinate system, see Fig.~\ref{fig:tri_vis} for a schematic visualisation. 
In the refinement stage, we store for every $\bfy^*_{j}$ the index of the voxel it belongs to, the indexes of the eight nearest displacements as well as the corresponding trilinear interpolation weights $\bfw \in \mathbb{R}^8$ in the point affinity table. 
The latter is then used in the backward pass of the refinement stage.

Let $x_{\text{max}}, y_{\text{max}}, z_{\text{max}}$ and $x_{\text{min}}, y_{\text{min}}, z_{\text{min}}$ be the maximum and minimum $x$-, $y$- and $z$-values among the eight nearest lattice point coordinates, respectively. 
To convert $\bfy_j^*$ from the coordinate system of the lattice to the local coordinate system, we calculate normalised distances $l_x$, $l_y$ and $l_z$: 
{\small
\begin{equation}
\hspace{-5pt}l_{ x }=\frac { x_j-x_{\text{min}} }{ x_{\text{max}}-x_{\text{min}} }, \; { l_{ y }=\frac { y_j-y_{\text{min}} }{ y_{\text{max}}-y_{\text{min}} }  } \,\;\text{and}\;\; l_{ z }=\frac { z_j-z_{\text{min}} }{ z_{\text{max}}-z_{\text{min}}}. 
\end{equation}} 
\hspace{-3pt}The individual displacement $\vec{\bfv}_{j}$ of $\bfy^*_j$ is obtained by trilinear interpolation of the eight nearest displacements, \textit{i.e.,} as an inner product of $\bfw$ and $x$-, $y$- and $z$-components of $\bfd$: 
{\small 
\begin{equation} 
\begin{split} 
\vec{\bfv}_{j, x}=\mathbf{w}^{\mathsf{T}} \mathbf{d_x} = 
\begin{bmatrix} (1-{ l }_{ x })(1-{ l }_{ y })(1-{ l }_{ z }) \\ (1-{ l }_{ x })(1-{ l }_{ y }){ l }_{ z } \\ (1-{ l }_{ x }){ l }_{ y }(1-{ l }_{ z }) \\ { l }_{ x }(1-{ l }_{ y })(1-{ l }_{ z }) \\  (1-{l }_{ x }){l}_{ y }{l }_{ z } \\ { l }_{ x }{ l }_{ y }(1-{ l }_{ z }) \\ { l }_{ x }(1-{ l }_{ y }){ l }_{ z } \\ { l }_{ x }{ l }_{ y }{ l }_{ z } \end{bmatrix}^{\mathsf{T}}\begin{bmatrix} { \bfd }_{ 000,x } \\ { \bfd }_{ 001,x } \\ { \bfd }_{ 010,x } \\ { \bfd }_{ 100,x } \\ { \bfd }_{ 011,x } \\ { \bfd }_{ 110,x } \\ { \bfd }_{ 101,x } \\ { \bfd }_{ 111,x } \end{bmatrix}, %
\end{split}
\end{equation}}
{\small 
\begin{equation} 
\begin{split} 
\vec{\bfv}_{j, y}=\mathbf{w}^{\mathsf{T}} \mathbf{d_y} = 
\begin{bmatrix} (1-{ l }_{ x })(1-{ l }_{ y })(1-{ l }_{ z }) \\ (1-{ l }_{ x })(1-{ l }_{ y }){ l }_{ z } \\ (1-{ l }_{ x }){ l }_{ y }(1-{ l }_{ z }) \\ { l }_{ x }(1-{ l }_{ y })(1-{ l }_{ z }) \\  (1-{l }_{ x }){l}_{ y }{l }_{ z } \\ { l }_{ x }{ l }_{ y }(1-{ l }_{ z }) \\ { l }_{ x }(1-{ l }_{ y }){ l }_{ z } \\ { l }_{ x }{ l }_{ y }{ l }_{ z } \end{bmatrix}^{\mathsf{T}}\begin{bmatrix} { \bfd }_{ 000,y } \\ { \bfd }_{ 001,y } \\ { \bfd }_{ 010,y } \\ { \bfd }_{ 100,y } \\ { \bfd }_{ 011,y } \\ { \bfd }_{ 110,y } \\ { \bfd }_{ 101,y } \\ { \bfd }_{ 111,y } \end{bmatrix}, %
\end{split}
\end{equation}}
and 
{\small 
\begin{equation} 
\begin{split} 
\vec{\bfv}_{j, z}=\mathbf{w}^{\mathsf{T}} \mathbf{d_z} = 
\begin{bmatrix} (1-{ l }_{ x })(1-{ l }_{ y })(1-{ l }_{ z }) \\ (1-{ l }_{ x })(1-{ l }_{ y }){ l }_{ z } \\ (1-{ l }_{ x }){ l }_{ y }(1-{ l }_{ z }) \\ { l }_{ x }(1-{ l }_{ y })(1-{ l }_{ z }) \\  (1-{l }_{ x }){l}_{ y }{l }_{ z } \\ { l }_{ x }{ l }_{ y }(1-{ l }_{ z }) \\ { l }_{ x }(1-{ l }_{ y }){ l }_{ z } \\ { l }_{ x }{ l }_{ y }{ l }_{ z } \end{bmatrix}^{\mathsf{T}}\begin{bmatrix} { \bfd }_{ 000,z } \\ { \bfd }_{ 001,z } \\ { \bfd }_{ 010,z } \\ { \bfd }_{ 100,z } \\ { \bfd }_{ 011,z } \\ { \bfd }_{ 110,z } \\ { \bfd }_{ 101,z } \\ { \bfd }_{ 111,z } \end{bmatrix}. %
\end{split}
\end{equation}}
Note that $\bfw$, $l_x$, $l_y$ and $l_z$ are shared across all dimensions.

\begin{figure}[t!]
 \centering
  \includegraphics[width=0.8\linewidth]{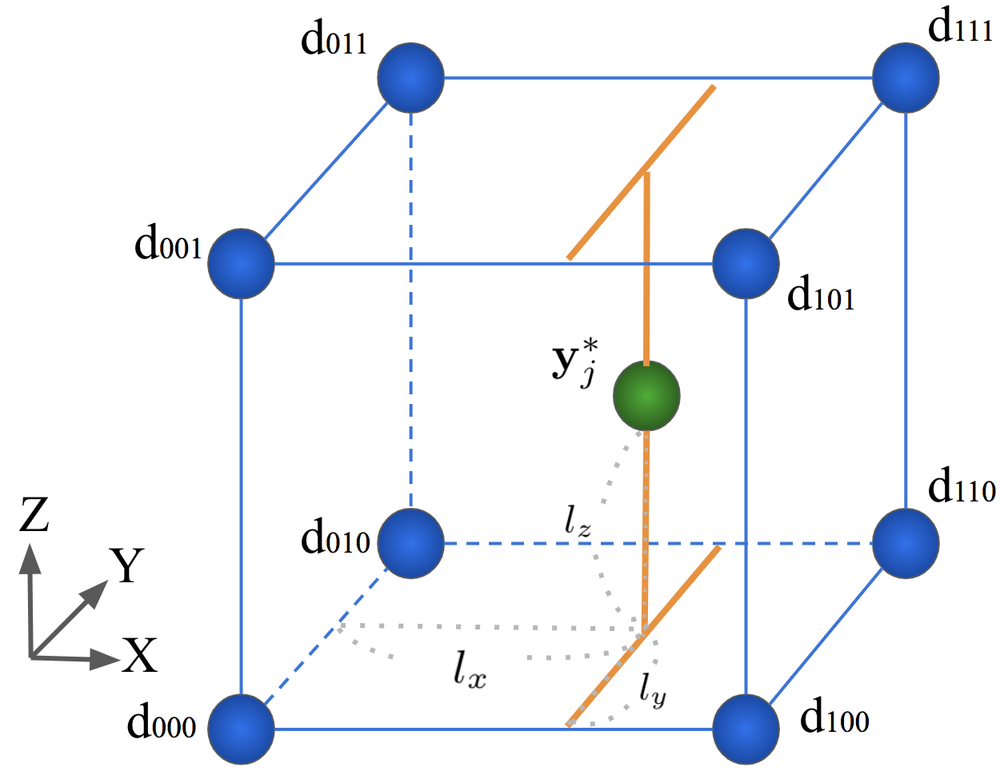}
  \caption{Schematic visualisation of trilinear interpolation for a given $\bfy^*_j$.} 
  \label{fig:tri_vis} 
\end{figure} 

\begin{figure*}[t!]
 \centering
  \includegraphics[width=0.9\linewidth]{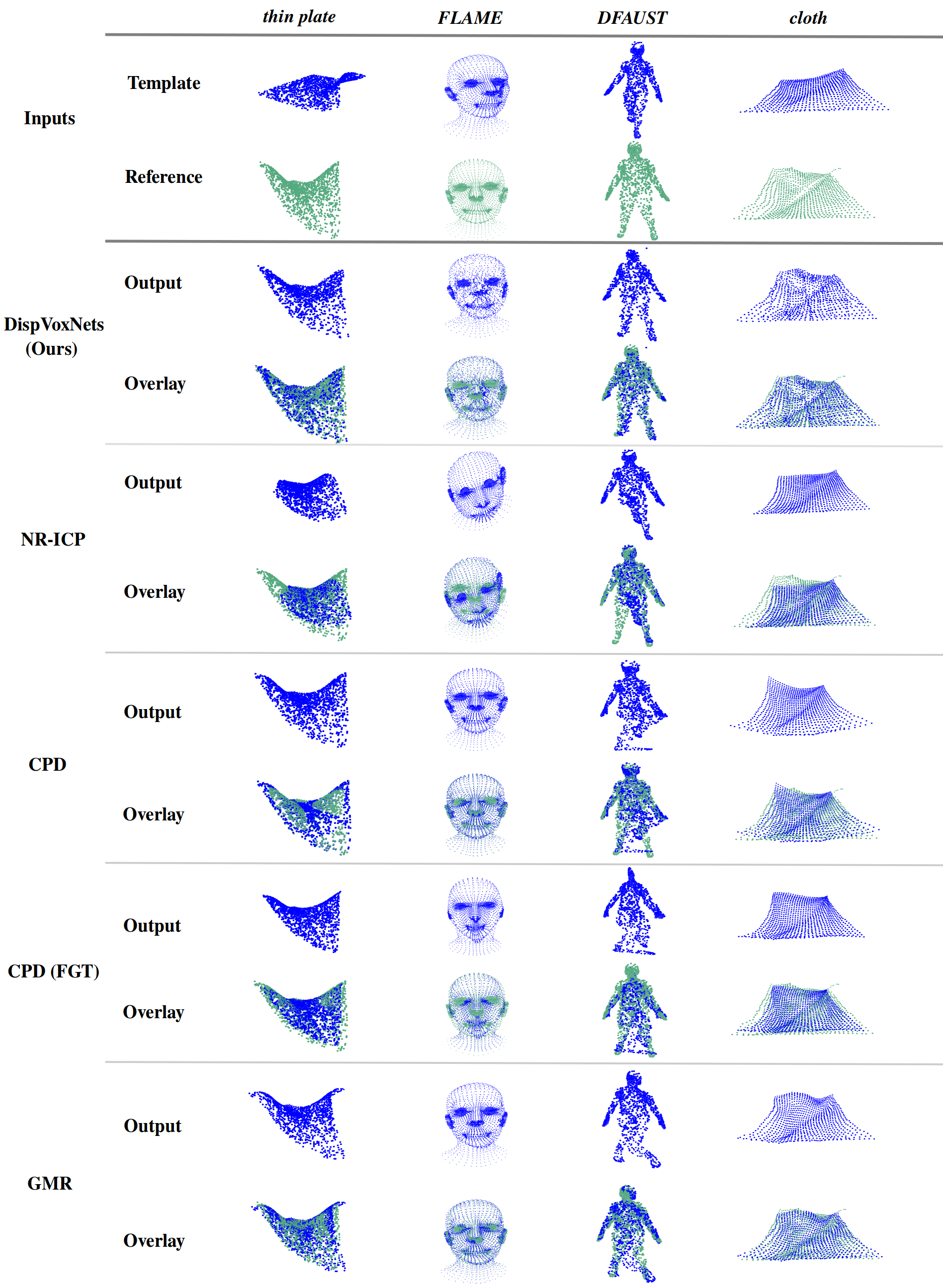} 
  \caption{Qualitative comparison of our DispVoxNets approach and other methods (NR-ICP \cite{Besl1992}, CPD/CPD with FGT \cite{Myronenko2010} and GMR \cite{Jian2005}). The input samples from each dataset are shown in the top rows, followed by the results (aligned templates and overlayed samples) for every method. } 
  \label{fig:qual_comp} 
\end{figure*}

\vspace{10pt} 

\subsection{Training Statistics}\label{sec:train_time} 
Table \ref{tab:iter} shows the number of training iterations until convergence for each dataset.
Since \textit{DFAUST} contains relatively large displacements between point sets, it requires the highest number of iterations followed by \textit{thin plate} and \textit{cloth}. 
On the contrary, \textit{FLAME} contains only small displacements, and the network requires fewer parameter updates to converge compared to other datasets.

\begin{table}
\center
 \scalebox{0.98}{
 \begin{tabular}{|c|c|c|c|c|}\hline
				    & \textit{\textbf{thin plate}} & \textit{\textbf{FLAME}} & \textit{\textbf{DFAUST}} & \textit{\textbf{cloth}}\\  \hline
      DE stage		    & 530k & 400k & 715k  & 500k \\ \hline
      refinement        & 14k		& 20k			& 24k &   12k  \\ \hline
\end{tabular}
}
\caption{\label{tab:iter} Number of training iterations for DispVoxNets in the DE and refinement stages for the tested datasets. 
}
\end{table}

\subsection{Qualitative Analysis and Observations} \label{sec:qual_comp} 

In this section, we provide additional qualitative results. 
In Fig.~\ref{fig:qual_comp}, we show selected registrations by our approach and other tested methods (NR-ICP \cite{Chui2003}, CPD/CPD with FGT \cite{Myronenko2010}, and GMR \cite{Jian2005}) on the tested datasets (\textit{thin plate}  \cite{Golyanik2018}, \textit{FLAME} \cite{FLAME:SiggraphAsia2017}, \textit{DFAUST} \cite{dfaust2017} and \textit{cloth} \cite{bednarik2018learning}). 

On the \textit{thin plate} --- due to the rather simple object structure --- all approaches except NR-ICP align the point sets reasonably accurate. 
CPD and DispVoxNets produce qualitatively similar results in the shown example. 
All methods show similar qualitative accuracy on the \textit{cloth} dataset, while differences are noticeable in the corners and areas with large wrinkles. 
At the same time, only our approach simultaneously captures both small and large wrinkles. 
Thus, many fine foldings present in the reference surface are not well recognisable in the aligned templates in the case of NR-ICP, CPD/CPD with FGT and GMR. %
All in all, results of these methods appear to be oversmoothed. 

In the absence of large displacements between the point sets --- which is the case with  \textit{FLAME} dataset --- model-based approaches CPD and GMR regress the displacements most accurately. 
The result of \hbox{DispVoxNets} is of comparable quality, though the deformed template is perceptually rougher and the points are arranged less regularly. %
This is due to the intermediate conversion steps from the point cloud representation to the voxel grid and back. 
We see that for small displacements, the limited resolution of the voxel grid 
is a more influential factor on the accuracy than the deformation prior learned from the data. 
With an increase of the voxel grid resolution, we expect our approach 
to come closer to CPD and GMR, up to the complete elimination of the accuracy gap (this is the matter of future work; currently, our focus is handling of large deformations which is a more challenging problem).

Next, we see that model-based approaches with global regularisers often fail on the \textit{FAUST} dataset, while the proposed approach demonstrates superior quantitative and visual accuracy. 
Even though the surface produced by \hbox{DispVoxNets} after the refinement stage can still seem coarse at some parts, 
the overall pose and shape are correctly and realistically inferred as we expect, despite substantial differences between the template and reference in the feet area (a subject standing on one foot and a subject standing on both feet respectively). 
Thus, model-based methods have difficulty in aligning the feet. 

Overall, the qualitative results in Fig.~\ref{fig:qual_comp} demonstrate the advantages of \hbox{DispVoxNets} for non-rigid point set alignment over classic, non-supervised learning-based approaches. 
Since our technique learns class-specific priors implicitly during training, it is successful in registering samples with large displacements and articulations.

\vspace{5pt} 

\subsection{Additional Experiments with Noisy Data}\label{sec:uni_noise} 

We present further experimental results with uniform noise in this section. 
Fig.~\ref{fig:uni_noise_S} shows RMSE graphs for various combinations of uniform noise ratios in the reference and template for all four datasets (\textit{thin plate}  \cite{Golyanik2018}, \textit{FLAME} \cite{FLAME:SiggraphAsia2017}, \textit{DFAUST} \cite{dfaust2017} and \textit{cloth} \cite{bednarik2018learning}). 

For previous methods, we observe the %
tendency that adding uniform noise to both the template and the reference can result in a lower error than only adding it to one of them. 
It is reasonable to assume that two point sets certainly differ more if noise is added to only one of them. 
Thus, when inputs contain a similar amount of noise (we can say that the noise levels  correlate), we observe the tendency that the alignment error becomes lower  (\textit{e.g.,}~see the seventh row, third column), \textit{i.e.,} some graphs roughly show a U-curve bottoming out at around $50\%$ of the added noise in the template. 
We hypothesise that this is due to what we call the \textit{mutual noise compensation effect}. 
Further study is required to clarify a more precise reason (it is possible that our observations are dataset-specific). 
Note that adding noise to both point sets is not a common evaluation setting. 
Usually, either template or reference is augmented with noise (\textit{cf.}~experimental sections in \cite{Jian2005, Myronenko2010, Ma2017, Ali2018}). 
With our experiment, we go beyond the prevalent evaluation methodology with noisy point sets.

On the one hand, CPD has the most stable error curve among the four model-based approaches, followed by NR-ICP and GMR. 
GMR shows higher errors when the noise is only in the template rather than in the reference, and CPD with FGT is the least stable as the noise ratio increases. 
Moreover, we observe that the relative performance of NR-ICP increases with the added noise. 
Thus, NR-ICP outperforms CPD only on \textit{DFAUST} according to Table~\ref{tab:base_comp} (the experiment with no added noise). 
In Fig.~\ref{fig:uni_noise_S}, we recognise multiple cases when NR-ICP outperforms CPD also on \textit{FLAME} (the blue curve is below the red curve).

On the other hand, our approach with \hbox{DispVoxNets} shows almost constant error  through all noise ratio combinations and all datasets. 
Compared to the case without noise, it even achieves the lowest RMSE on \textit{FLAME} for multiple noise level combinations (${\sim}40\%$ of the cases). 
As our network becomes aware of class-specific features after the training and learns to ignore noise, 
it can distinguish the meaningful shapes from noise, which contributes to its overall robustness. 
To the best of our belief, it is for the first time that a NRPSR method is so stable, even in the presence of large amount of noise in the data. 
Recall that we follow a simple noise augmentation policy for the training data (Sec.~\ref{sec:training_detail}). 
Thus, our framework seemingly learns to filter uniform noise. 
Another factor could be that individual unstructured points cause neuron deactivations.  
In future work, it could be interesting to study augmentation policies for further types of noise (\textit{e.g.,} noise along the surfaces).

\begin{figure*}[t]
 \centering
  \includegraphics[width=1.0\linewidth]{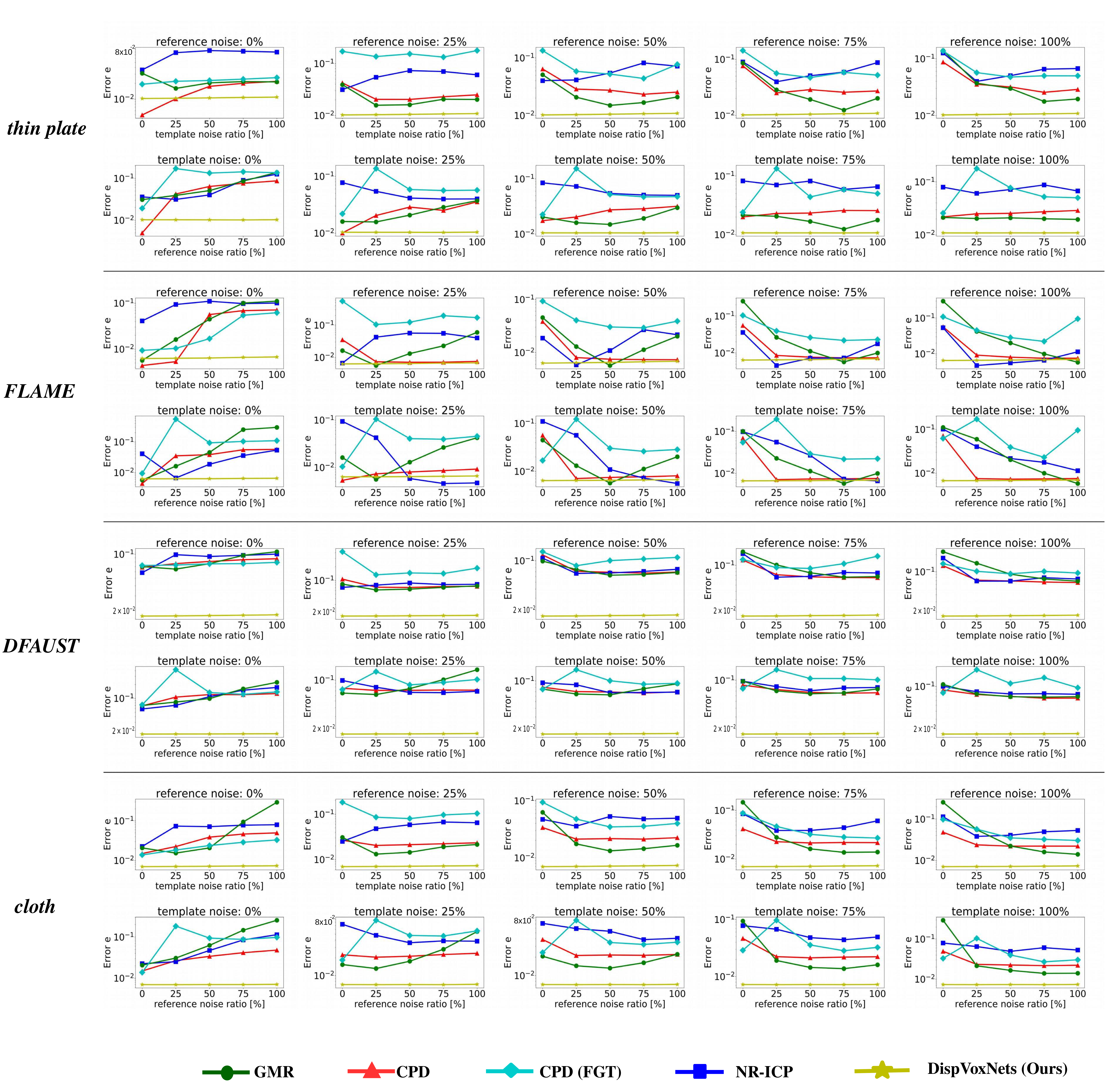} 
  \caption{RMSE ($e$) graphs for additional experiments with uniform noise 
  on \textit{thin plate} \cite{Golyanik2018}, \textit{FLAME} \cite{FLAME:SiggraphAsia2017}, \textit{DFAUST} \cite{dfaust2017} and \textit{cloth} \cite{bednarik2018learning} datasets. 
  $p\%$ is the ratio between the number of added points and the number of points in the sample. 
  In this experiment, both reference and template are augmented with noise. 
  } 
  \label{fig:uni_noise_S} 
\end{figure*} 

\end{alphasection}

\end{document}